\documentclass{article}

\PassOptionsToPackage{sort&compress, comma, square, numbers}{natbib}

\usepackage[final]{neurips_2019}

\usepackage[utf8]{inputenc} 
\usepackage[T1]{fontenc}    
\usepackage{hyperref}       
\usepackage{url}            
\usepackage{booktabs}       
\usepackage{amsfonts}       
\usepackage{nicefrac}       
\usepackage{microtype}      

\usepackage{subcaption}
\usepackage{wrapfig}
\usepackage{times}
\usepackage{epsfig}
\usepackage{graphicx}
\usepackage{amsmath}
\usepackage{amssymb}
\usepackage{pifont}
\usepackage{booktabs}

\usepackage{lineno}
\modulolinenumbers[5]
\usepackage{graphicx}
\usepackage{booktabs}
\usepackage{amssymb,amsmath,nccmath}
\usepackage{cclicenses}
\usepackage{makecell}
\usepackage{lscape,array}
\newcolumntype{C}[1]{>{\centering\arraybackslash}p{#1}} 
\usepackage[thin, , thinc]{esdiff}
\usepackage{subcaption}
\usepackage{caption}
\usepackage{framed}  
\usepackage[font=small,skip=0pt]{caption}
\usepackage{color}

\title{A Self Validation Network for Object-Level Human Attention Estimation}

\author{%
  Zehua Zhang,$^1$ Chen Yu,$^2$ David Crandall$^1$ \\
  $^1$Luddy School of Informatics, Computing, and Engineering\\
 $^2$Department of Psychological and Brain Sciences\\
  Indiana University Bloomington\\
  \texttt{\{zehzhang, chenyu, djcran\}@indiana.edu} \\
}

\begin{document}

\maketitle

\begin{abstract}
Due to the foveated nature of the human vision system, people can
focus their visual attention on only a small region of their visual
field at a time, which usually contains a single object.  Estimating
this object of attention in first-person (egocentric) videos is useful
for many human-centered real-world applications such as augmented
reality and driver assistance systems. A straightforward
solution for this problem is to first estimate the gaze with a traditional gaze estimator 
and generate object candidates from an off-the-shelf object detector, and then pick the object that the estimated gaze falls on.
However, such an approach can fail because it addresses the
 \textit{where} and the \textit{what}
  problems separately, 
despite that they are highly related, chicken-and-egg
problems. In this paper, we propose a novel unified model that
incorporates both spatial and temporal evidence in identifying as well as locating
the attended object in first-person videos. It introduces a novel
Self Validation Module that enforces and leverages consistency of the
\emph{where} and the \emph{{what}} concepts. We
evaluate on two public datasets, demonstrating
that the Self Validation Module significantly benefits both training
and testing and that our model outperforms the state-of-the-art. 
\end{abstract}

\section{Introduction}

\begin{wrapfigure}{R}{0.5\textwidth} 
\begin{center}
\begin{subfigure}{.25\textwidth}
  \centering
  \includegraphics[width=.9\linewidth]{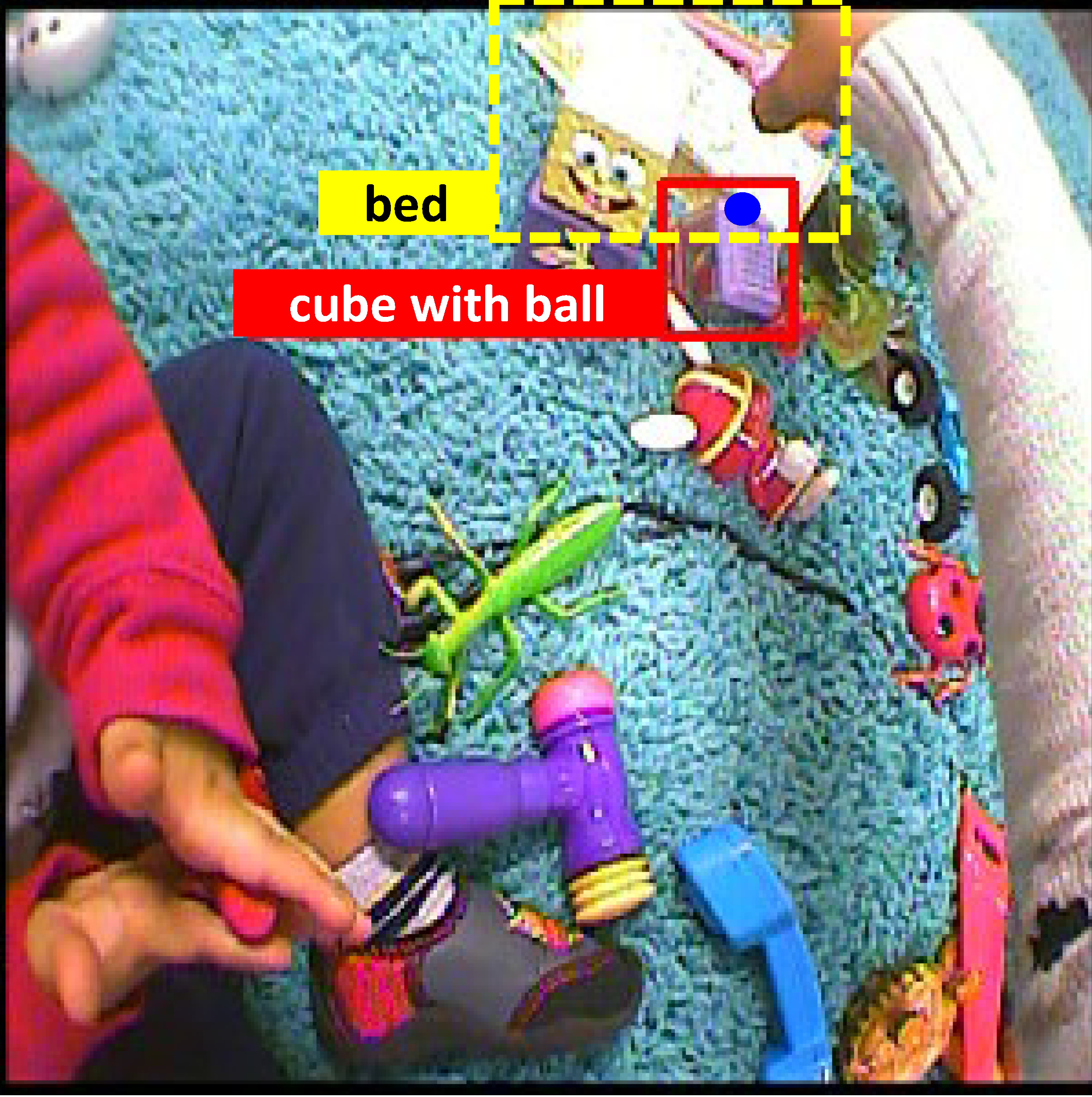}
  \caption{}
  \label{fig:introsfig1}
\end{subfigure}%
\begin{subfigure}{.25\textwidth}
  \centering
  \includegraphics[width=.9\linewidth]{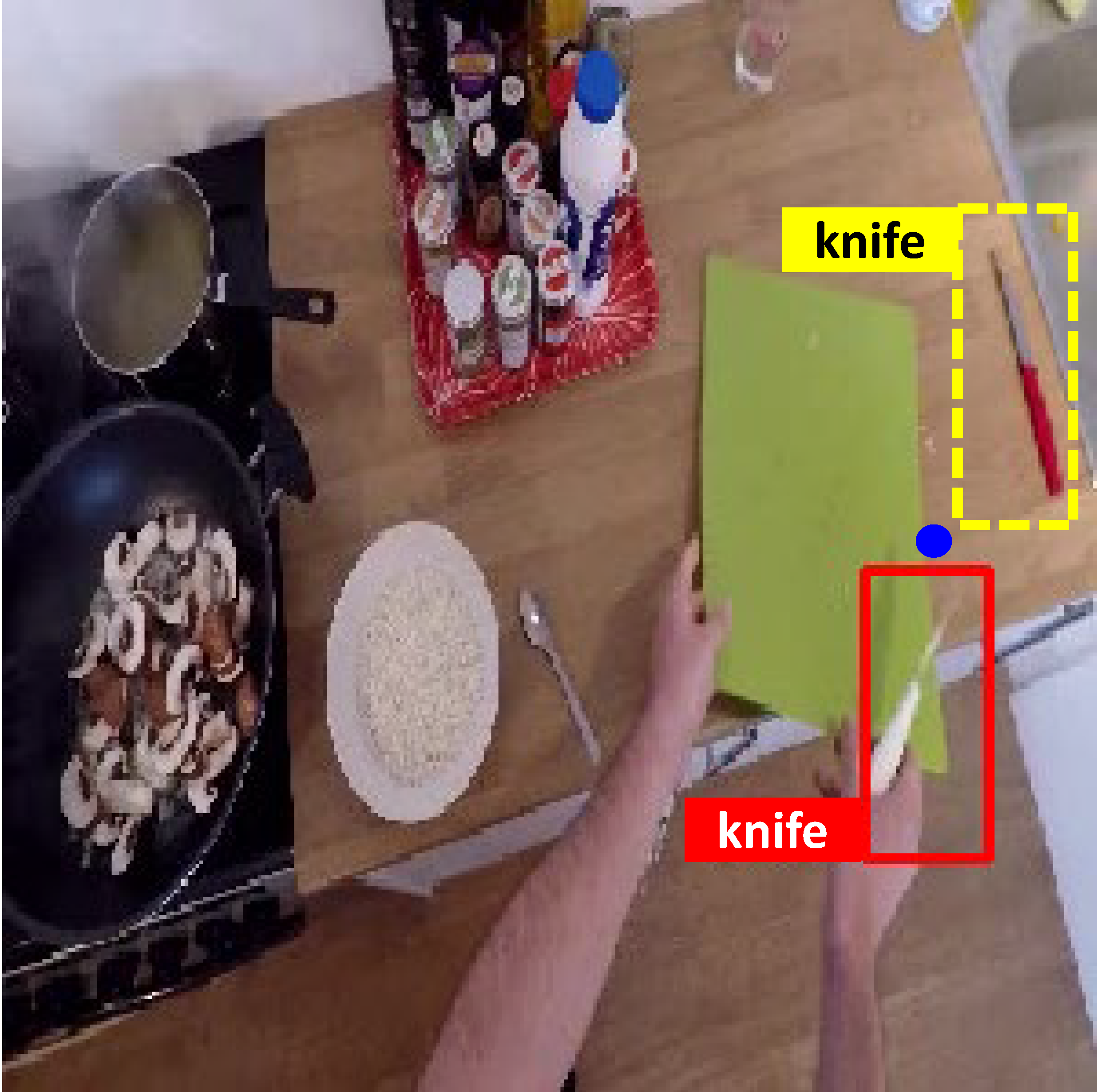}
  \caption{}
  \label{fig:introsfig2}
\end{subfigure}
\end{center}
  \caption{\textit{Among the many objects appearing in an egocentric video frame of  a person's
      field of view, we want to
      identify and locate the object to which the person is visually attending.}
    Combining traditional eye gaze estimators and existing object detectors
can fail when eye gaze prediction (blue dot) is
    slightly incorrect, such as when (a) it falls in the intersection
    of two object bounding boxes or (b) it lies between two bounding
    boxes sharing the same class. Red boxes shown actual attended
    object according to ground truth gaze and yellow dashed boxes show
    incorrect predictions.}
\label{fig:intro}
\end{wrapfigure}

Humans can 
focus their visual attention on only a small part of their surroundings at any moment, 
and thus have to choose what
to pay attention to in real time~\cite{mozer1998computational}. 
Driven by the tasks and intentions we have in mind,
we manage attention with our foveated visual system by
adjusting our head pose and our gaze point in order to 
focus on the  most relevant object in the environment at any moment in time~\cite{lazzari2009eye, bowman2009eye, hayhoe2005eye,
  vidoni2009manual, perone2008relation}.

This close relationship between intention, attention, and semantic
objects has inspired a variety of work in computer vision, including image classification~\cite{karessli2017gaze},
object detection~\cite{papadopoulos2014training, karthikeyan2013and,
  shcherbatyi2015gazedpm, rutishauser2004bottom}, action
recognition~\cite{li2018eye, baradel2018object, adl, ma2016going},
action prediction~\cite{shen2018egocentric}, 
video summarization~\cite{lee2012discovering}, visual search
modeling~\cite{sattar2015prediction}, and irrelevant frame
removal~\cite{liu2010hierarchical}, in which the attended object estimation serves as auxiliary information. Despite being a key component of these
papers,  how to identify and locate the important object is seldom studied
explicitly. This problem in and of itself is  of broad potential use
in real-world applications such as driver assistance systems
and intelligent human-like robots.
%
%
%

In this paper, we discuss how to identify and locate the attended
object in first-person videos. Recorded by head-mounted cameras along with eye trackers, first-person videos capture an approximation of what people see in their
fields of view as they go about their lives, yielding interesting data for studying real-time human attention. In contrast to gaze studies
of static images or pre-recorded videos, first-person video is unique
in that there is exactly one correct point of attention in each
frame, as a camera wearer can only gaze at one point at a time. Accordingly, one and only one gazed object exists for each frame, reflecting the camera wearer’s real-time attention and intention. We will use the term~\textbf{\emph{object of interest}} to refer to the attended object in our later discussion. 

Some recent work~\cite{t3f,huang2018predicting,dfg2017cvpr}
has discussed estimating probability maps of ego-attention or predicting gaze points in egocentric videos. However, people think not in terms of points in their field of
view, but in terms of the \textit{objects} that they are attending to. Of course, the object of interest could be obtained by first estimating the gaze with the gaze estimator and generating object candidates from an off-the-shelf object detector, and then picking the object that the estimated gaze falls in. Because this bottom-up approach estimates \textit{where} and \textit{what} separately, it could be doomed to fail if the
eye gaze prediction is slightly inaccurate, such as falling between two
objects or in the intersection of multiple object bounding boxes
(Figure~\ref{fig:intro}). To assure consistency, one may think of performing anchor-level attention estimation and directly predicting the attended box by modifying existing object detectors. Class can be either predicted simultaneously with the anchor-level attention estimation using the same set of features, as in SSD~\cite{ssd}, or afterwards using the features pooled within the attended box, as in Faster-RCNN~\cite{fasterrcnn}. Either way, these methods still do not yield satisfying performance, as we will show in Sec.~\ref{sec:resatt}, because they lack the ability to leverage the consistency to refine the results.  




We propose to identify and locate the object of interest by \textit{jointly}
estimating \textit{where} it is within the frame as
well as recognizing \textit{what} its identity is. 
In
particular, we propose a novel model --- which we cheekily call
Mindreader Net or Mr. Net --- to jointly solve the
problem. Our model incorporates both spatial evidence within frames
and temporal evidence across frames, in a network architecture (which
we call the Cogged Spatial-Temporal Module) with separate spatial and
temporal branches to avoid feature entanglement.  

A key feature of our model is that it  explicitly enforces and leverages a simple but
extremely useful constraint: our estimate of \textit{what} is being attended should be located
in exactly the position of \textit{where} we estimate the attention to be.
This Self Validation Module  first computes similarities between the global
object of interest class prediction vector and each local anchor box
class prediction vector as the attention validation score to update
the anchor attention score prediction, and then, with the updated anchor
attention score, we select the attended anchor and use its
corresponding class prediction score to update the global object of
interest class prediction. With global context originally incorporated by extracting features from the whole clip using 3D convolution, the Self Validation Module
helps the network focus on the local context in a spatially-local
anchor box and a temporally-local frame. 

We evaluate the approach on two existing first-person video datasets
that include attended object ground truth annotations. We show our
approach outperforms baselines, and that our Self Validation Module not only improves performance by refining the outputs
with visual consistency during testing, but also 
it helps bridge multiple components together during training
to guide the model to learn a highly meaningful latent representation.
More information is available
at~\url{http://vision.soic.indiana.edu/mindreader/}.

\section{Related Work}
\label{sec:iod}

Compared with many efforts to understand human attention by modeling
eye gaze~\cite{t3f, dfg2017cvpr, huang2018predicting, li2014secrets,
  gteaplusyinli, ittiBottomUp, graphbased, spectralbased, salicon,
  nn2016Pan, integrationgaze, Torralba06contextualguidance, baeyegaze,
  twostream1borji, eyegazeYamada} or
saliency~\cite{sal1,sal2,sal4,sal5,sal6,sal7,sal8,sal9,sal10}, there
are relatively few papers that  detect object-level attention.
Lee~\textit{et al.}~\cite{lee2012discovering} address video
summarization with hand-crafted features to detect important
people and objects, while object-level reasoning plays a key role in
Baradel~\textit{et al.}'s work on understanding videos through interactions of
important objects~\cite{baradel2018object}. 
In the particular case of egocentric video, Pirsiavash and
Ramanan~\cite{adl} and Ma~\textit{et al.}~\cite{ma2016going} detect
objects in hands as a proxy for attended objects to help action
recognition. 
However, 
eye gaze usually precedes hand motion and thus objects in hand 
are not always those being visually attended
(Fig.~\ref{fig:introsfig1}). Shen~\textit{et
  al.}~\cite{shen2018egocentric} combine eye gaze ground
truth and detected object bounding boxes to extract attended object
information for future action prediction. 
EgoNet~\cite{bertasius2016first}, among the
first papers to focus on important object detection in
first-person videos, combines visual appearance and
3D layout information to generate probability maps of object
importance. Multiple objects
 can be detected in a single frame, 
making their results more similar to saliency than human
attention in egocentric videos.

Perhaps the most related work to ours is
Bertasius~\textit{et al.}'s Visual-Spatial Network (VSN)~\cite{bertasius2017unsupervised}, which
proposes an unsupervised method for important object detection in
first-person videos that incorporates the idea of
consistency between the \textit{where} and \textit{what}
concepts to facilitate learning. However, VSN requires a much more
complicated training strategy of switching the cascade order of the
two pathways multiple times, whereas we present a unified framework that
can be learned end-to-end.

\vspace{-8pt}

\section{Our approach}
\begin{figure*}[t]
    \begin{center}
       \includegraphics[width=1.0\textwidth]{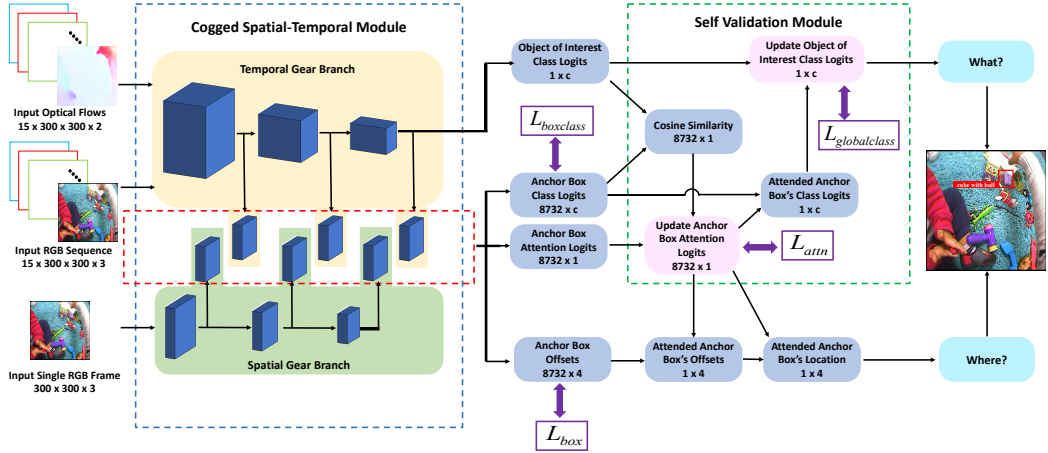}
    \end{center}
\caption{\emph{The architecture of our proposed
    Mindreader Net.} Numbers indicate
    output size of each component (where $c$ is the number of object classes). 
Softmax is applied before computing 
    the losses on global classification  $L_{globalclass}$, anchor box
    classification  $L_{boxclass}$, and attention 
    $L_{attn}$ (which is first
    flattened to be 8732-d). Please refer  to 
    supplementary materials for details about the Cogged
    Spatial-Temporal Module.}  \label{fig:modelVis} \vspace{-12pt}
\end{figure*}

Given a video captured with a head-mounted camera, our goal is to
detect the object that is visually attended in each frame. This is
challenging because egocentric videos can be highly cluttered, with
many competing objects vying for attention.  We thus incorporate
temporal cues that consider multiple frames at a time.  We first
consider performing detection for the middle frame of a short input sequence (as
in~\cite{ma2016going}), and then further develop it to work online
(considering only past information) by performing detection on the
last frame.  Our novel model consists of two main parts (Figure~\ref{fig:modelVis}), which we call the Cogged
Spatial-Temporal Module and the Self Validation Module.

\subsection{Cogged Spatial-Temporal Module}
The Cogged Spatial-Temporal Module consists of a spatial  and a
temporal branch. The ``cogs'' refer to the way that the outputs 
of each layer of the two branches are combined together, reminiscent of
the interlocking cogs of two gears (Figure~\ref{fig:modelVis}).
Please see supplementary material for more details.

\textbf{The Spatial Gear Branch,} inspired by SSD300~\cite{ssd},
takes a single frame $I_t$ of size $h\times w$ and performs spatial
prediction of local anchor box offsets and anchor box classes.  It is
expected to work as an object detector, although we only have ground
truth for the objects of interest to train it, so we
do not add an extra background class as in~\cite{ssd}, and only compute
losses for the spatial-based tasks on the matched positive anchors.
We use atrous~\cite{chen2018deeplab, yu2015multi}
VGG16~\cite{simonyan2014very} as the backbone and follow a similar
anchor box setting as~\cite{ssd}. We also apply the same
multi-anchor matching strategy. With the spatial branch, we obtain
anchor box offset predictions $O \in R^{a \times 4}$ and 
class predictions $C_{box} \in R^{a \times c}$, where $a$ is the number
of anchor boxes and $c$ is the number of classes in our
problem. Following SSD300~\cite{ssd}, we have $a=8732$, $h=300$, and
$w=300$.

\textbf{The Temporal Gear Branch} takes $N$ continuous 
RGB frames $I_{t-\frac{N-1}{2},t+\frac{N-1}{2}}$ as well as $N$
corresponding optical flow fields
$F_{t-\frac{N-1}{2},t+\frac{N-1}{2}}$, both of spatial resolution
$h \times w$ (with $N=15$, set empirically). We use
Inception-V1~\cite{szegedy2015inception} I3D~\cite{i3d} as the
backbone of our temporal branch.
With aggregated global features from 3D
convolution, we obtain global object of interest class predictions
$C_{global} \in R^{1 \times c}$ and anchor box attention predictions
$A \in R^{a \times 1}$. We match the ground truth box only to the
anchor with the greatest overlap (intersection over union). The
matching strategy is empirical and discussed in
Section~\ref{sec:onehot}.

\subsection{Self Validation Module}\label{sec:mrm}
The Self Validation Module connects the above
branches and delivers global and local context between the two
branches at both spatial (\emph{e.g.}, whole frame versus an
anchor box) and temporal (\emph{e.g.}, whole
sequence versus a single frame) levels. It incorporates the constraint on consistency
between where and what by embedding a double
validation mechanism: what$\xrightarrow{}$where and
where$\xrightarrow{}$what.

\textbf{What$\xrightarrow{}$where.} With the 
outputs of the Cogged Spatial-Temporal Module, we compute the cosine
similarities between the global class prediction 
$C_{global}$ and the class prediction for each anchor box,
$C_{box_i}$, yielding an attention validation score for each box $i$,
\begin{equation}
    V^i_{attn}=\frac{C_{global} C^T_{box_i}}{||C_{global}|| \times ||C_{box_i}||.}
\end{equation}
Then the attention validation vector $V_{attn} = [V^1_{attn},
V^2_{attn}, ..., V^a_{attn}] \in R^{a \times 1}$ is used to update the
anchor box attention scores $A$ by element-wise summation, $A' = A +
V_{attn}$. Since $-1 \leq V^i_{attn} \leq 1$, we make
the optimization easier by rescaling each $A_i$ 
to the range  $[-1,1]$,
%
%
\begin{equation}
~\label{eq:val1}
    A' = R(A) + V_{attn}= \frac{A - (max(A) + min(A))/2}{max(A) - (max(A) + min(A))/2} + V_{attn},
\end{equation}
where $max()$ and $min()$ are element-wise vector operations.

\textbf{Where$\xrightarrow{}$what.} 
Intuitively, 
obtaining the attended anchor box index $m$ is a simple matter of
computing
$m=argmax(A')$, and the class validation score is
simply
$V_{class}=C_{box_{m}}$. Similarly, after rescaling, we take an element-wise
summation, $V_{class}$ and $C_{global}$, to update the global object of interest class
prediction ($R(\cdot)$ in Equation~\ref{eq:val1}), $C_{global}' = R(C_{global}) + R(V_{class})$.
However, the hard argmax is not differentiable, and thus 
gradients are not able to backpropagate properly during training. 
We thus use soft argmax.
Softmax is applied to the updated
anchor box attention score $A'$ to produce a weighting vector
$\widetilde{A}'$ for class validation score estimation,
\begin{equation}
    \hat{V}_{class} = \sum_{i=1}^a \widetilde{A}_i'C_{box_{i}}, ~~~~~~~~~~~~~\mbox{with}~~ \widetilde{A}_i' = \frac{e^{A_i'}}{\sum_{j=1}^{a}e^{A_j'}}
\end{equation}
Now we replace $V_{class}$ with $\hat{V}_{class}$ to update $C_{global}$,    $C_{global}' = R(C_{global}) + R(\hat{V}_{class})$.

This
soft what$\xrightarrow{}$where
validation is closely related to the soft attention mechanism widely
used in many recent papers~\cite{sukhbaatar2015end, bahdanau2014neural,
cho2015describing, soft1, soft2, vaswani2017attention}. While soft
attention learns the mapping itself inside the model, we explicitly
incorporate the coherence of the where and what concepts into our
model to self-validate the output during both training and
testing. In contrast to  soft attention which describes relationships
between \textit{e.g.} words, graph
nodes, \textit{etc.}, this self-validation mechanism naturally
mirrors the visual consistency of our foveated vision system.

\subsection{Implementation and training details}
We implemented our model with Keras~\cite{keras} and
Tensorflow~\cite{tensorflow}. A batch normalization
layer~\cite{batchnorm} is inserted after each layer in both spatial
and temporal backbones, and momentum for batch normalization
is $0.8$. Batch normalization is not used in the four prediction heads.
We found pretraining the spatial branch helps the model converge
faster. No extra data is introduced as we still only use the labels of
the objects of interest for pretraining. 
VGG16~\cite{simonyan2014very} is initialized with weights
pretrained on ImageNet~\cite{deng2009imagenet}.
We use Sun~\textit{et al.}'s method~\cite{Sun2018PWC-Net, pytorch-pwc} to
extract optical flow and follow~\cite{i3d} to truncate the maps
to $[-20, 20]$ and then rescale them to $[-1, 1]$. The
RGB input to the Temporal Gear Branch is rescaled to $[-1,
1]$~\cite{i3d}, while for the Spatial Gear Branch  the RGB input 
is normalized to have 0 mean  and the channels are permuted
to BGR.

When training the whole model, the spatial branch is initialized with
the pretrained weights from above. The I3D backbone is initialized
with weights pretrained on Kinetics~\cite{kay2017kinetics} and
ImageNet~\cite{deng2009imagenet}, while other parts are randomly
initialized. We use stochastic gradient descent with learning rate 
$0.03$, momentum $0.9$, decay $0.0001$, and $L2$ regularizer
$5e^{-5}$. The loss function consists of four parts:
global classification  $L_{globalclass}$, attention 
$L_{attn}$, anchor box classification  $L_{boxclass}$, and 
box regression  $L_{box}$,
\begin{equation}
    L_{total} = \alpha L_{globalclass} + \beta L_{attn} + \frac{1}{N_{pos}}(\gamma L_{boxclass} + L_{box}),
    \label{eq:totalloss}
\end{equation}
where we empirically set $\alpha=\beta=\gamma=1$, and $N_{pos}$ is the
total number of matched anchors for training the anchor box class
predictor and anchor box offset predictor. $L_{globalclass}$ and
$L_{attn}$ apply cross entropy loss, computed on the updated
predictions of object of interest class and anchor box
attention. $L_{boxclass}$ is the total cross entropy loss and
$L_{box}$ is the total box regression loss over only all the matched
anchors. The box regression loss follows~\cite{fasterrcnn, ssd}
and we refer readers there for details. Our full model has $64M$ trainable
parameters, while the Self Validation Module contains no parameters,
making it very flexible so that it can be added to training or testing
anytime. It is even possible to stack multiple Self Validation Modules
or  use  only half of it.

During testing, the anchor with the highest anchor box attention score
$A'_i$ is selected as the attended anchor. The corresponding
anchor box offset prediction $O_i$ indicates where the object of
interest is, while the argmax of the
global object of interest class score $C'_{global}$ gives its class.

\section{Experiments}

We evaluate our model on identifying attended objects in two first-person datasets
collected in 
very different contexts: child and adult toy play,
and adults in kitchens. 

\textbf{ATT~\cite{t3f}} (Adult-Toddler Toy play) consists of
first-person videos from head-mounted cameras of parents and toddlers
playing with 24 toys in a simulated home environment. The dataset
consists of 20 synchornized video pairs (child head cameras
and parent head cameras), although we only use the parent videos.
The object being attended is determined using 
gaze tracking. 
We randomly select $90\%$ of the samples in each
object class for training  and use the remaining $10\%$ 
for testing, resulting in about $17,000$ training and $1,900$
testing samples, each with 15 continuous frames. We do not restrict the object of interest to remain
the same in each sample sequence and only use the label of  the object
of interest for training.

\textbf{Epic-Kitchen Dataset~\cite{epickit}} contains $55$ hours of
first-person video from 32 participants in their
own kitchens. 
The dataset  includes
anntoations on the ``active'' objects related to the person's current action.
We use this as a proxy for attended object by
we selecting only frames containing one active object and assuming that they are attended.
Object classes with fewer than
$1000$ samples are also excluded, resulting in $53$ classes. We
randomly select 90\% of samples for training, yielding
about $120,000$ training 
and $13,000$ testing samples.

For evaluation, we report accuracy --- number of correct predictions
over the number of samples. A prediction is considered correct if it
has both (a) the correct class prediction and (b) an IoU between the
estimated and the ground truth boxes above a threshold. Similar
to~\cite{coco}, we report accuracies 
at IOU thresholds of $0.5$ and $0.75$, as well as a mean accuracy
$mAcc$ computed by averaging accuracies at $10$
IOU thresholds  evenly distributed from $0.5$ to $0.95$. 
Accuracy thus measures ability to correctly predict both
what and where is being attended.

\subsection{Baselines}

We evaluate against several strong baselines.
\textbf{Gaze + GT bounding box,} inspired by
Li~\textit{et al.}~\cite{li2014secrets}, 
applies
Zhang~\textit{et al.}'s gaze prediction method~\cite{t3f} (since it has state-of-the-art performance on the \emph{ATT})
and directly uses ground truth object bounding boxes.
This is
equivalent to having a perfect object detector (with $mAP=100\%$),
resulting in a very strong baseline. We use two different methods to
match the predicted eye gaze to the object 
boxes: (1) \textbf{Hit:} only boxes in which the gaze falls in are considered 
matched, and if the estimated gaze point is within multiple boxes, 
the accuracy score is averaged by the number of matched boxes; and
(2) \textbf{Closest:} the box whose center is the closest to the predicted
gaze is considered to be matched.
\textbf{I3D~\cite{i3d}-based SSD~\cite{ssd}} tries to overcome the
discrepancy caused by solving the \emph{where} and \emph{what} problems
separately by directly performing anchor-level attention estimation with an
I3D~\cite{i3d}-backboned SSD~\cite{ssd}. The anchor box
setting is similar to SSD300~\cite{ssd}. For each anchor we predict an
attention score, a class score, and box offsets.
\textbf{Cascade model} contains a temporal branch with
I3D backbone and a spatial branch with VGG16 backbone. From
the temporal branch, the important anchor as well as its box offsets
are predicted, and then features are pooled~\cite{fasterrcnn, he2017mask}
from the spatial branch for classification.
\textbf{Object in hands + GT bounding box,} inspired by~\cite{adl,
  ma2016going, nextactive}, tries to detect object of interest by
detecting the object in hand. We use several variants; the ``either
handed model'' is strongest, and uses both the ground truth object
boxes  and the ground truth label of the object in hands. When
two hands hold different objects, the model always picks the one yielding
higher accuracy, thus reflecting the best performance we
can obtain with this baseline. Please
refer to the supplementary materials for details of other variants.
\textbf{Center GT box} uses
the ground truth object boxes and labels to select the object
closest to the frame center, 
inspired by the fact that people tend to adjust
their head pose so that their gaze is near the center of
their view~\cite{gteaplusyinli}.

\subsection{Results on ATT dataset}
~\label{sec:resatt}
\begin{table}[t]
\begin{minipage}[t][][b]{.46\linewidth}
\scalebox{0.65}{
\begin{tabular}{@{}l c c c c @{}} 
\toprule
        Method  & $Acc_{0.5}~\uparrow$ & $Acc_{0.75}~\uparrow$ & $mAcc~\uparrow$ \\ \midrule
        Our Mr. Net   & \textbf{74.27} & \textbf{46.78} & \textbf{44.78}\\ 
\midrule
        Gaze~\cite{t3f} + GT Box + Hit   & 25.26 & 25.26 & 25.26\\ 
        Gaze~\cite{t3f} + GT Box + Closest   & 35.86 & 35.86 & 35.86\\
        I3D~\cite{i3d}-based SSD~\cite{ssd} & \textbf{70.11} & 42.10 & 40.85 \\
        Cascade Model & 66.97 & \textbf{45.10} & 41.93 \\
        OIH Detectors + WH Classifier & 37.16 & 37.16 & 37.16 \\
        Left Handed Model & 38.31 & 38.31 & 38.31 \\
        Right Handed Model & 39.00 & 39.00 & 39.00 \\
        OIH GT + WH Classifier  & 40.83 & 40.83 & 40.83 \\
        Either Handed Model & 42.94 & 42.94 & \textbf{42.94} \\
        Center GT Box & 23.97 & 23.97 & 23.97 \\
        \bottomrule
\end{tabular}}
\vspace{6pt}
          \caption{\textit{Accuracy of our method compared to
others,} on the ATT dataset. OIH represents Object-in-Hand, while WH means Which-Hand.}
      \label{tab:att}
\end{minipage}
\hspace{12pt}
\begin{minipage}[t][][b]{.49\linewidth}
    \centering
\scalebox{0.68}{
\begin{tabular}{@{}c c c c c c c @{} } 
\toprule
& \multicolumn{2}{c}{Self validation? } \\
\cmidrule{2-3}
Streams &         Training & Testing   & $Acc_{0.5}~\uparrow$ & $Acc_{0.75}~\uparrow$ & $mAcc~\uparrow$ \\ \midrule
      Two &  yes & yes    & \textbf{74.27} & \textbf{46.78} & \textbf{44.78}\\ 
\midrule
Two &        yes & half   & --- & --- & 43.88\\
Two &        yes & no   & 68.19 & 42.83 & 41.18\\
     Two &    no & yes    & 67.18 & 40.06 & 39.48 \\
   Two &      no &  half   & --- & --- & 37.87\\
    Two &     no & no   & 62.33 & 38.31 & 37.18 \\
    RGB &    yes & yes   & 74.59 & 43.15 & 42.48 \\
    Flow &    yes & yes   & 64.30 & 38.63 & 37.60 \\
   Flow &     no & yes   & --- & --- & 25.10 \\
   Flow &     no & no   & --- & --- & 18.40 \\
        \bottomrule
\end{tabular}}
\vspace{6pt}
      \caption{\textit{Ablation  results.} Testing with half means that the model is tested with only what$\xrightarrow{}$where validation.}
      \label{tab:ablation}
\end{minipage}
\vspace{-20pt}
\end{table}

Table~\ref{tab:att} presents quantitative results of 
our Mindreader Net and baselines on the \emph{ATT}
dataset. 
%
Both enforcing and leveraging the visual consistency, our method even outperformed the either-handed model in terms of $mAcc$, which
is built upon several strong \textit{oracles} --- a perfect
object detector, two perfect object-in-hand detectors, and a perfect
which-hand classifier. 
Other methods without perfect object
detectors suffer from a rapid drop in $Acc$ as the IOU
threshold becomes higher. 
For example, when the IOU threshold
reaches 0.75, the either-handed model already has no obvious
advantage compared with I3D-based SSD, and the Cascade
model achieves a much higher score. When the threshold becomes 0.5,
not only our Mindreader Net but also Cascade  and I3D-based SSD
outperform the either-handed model by a significant margin. Though the
$Acc_{0.5}$ of the cascade model is lower than I3D-based
SSD  by about $3\%$, its $mAcc$ and $Acc_{0.75}$ are higher,
suggesting bad box predictions with low IOU confuses the class head
of the cascade model, but having a separate spatial branch to overcome
feature entanglement improves the overall performance with
higher-quality predictions. 

We also observed that 
the Closest variant of the Gaze +
GT Box model 
is about $40\%$ better than the Hit variant. This suggests that gaze
prediction often misses the ground truth box a bit or may
fall in the intersection of several bounding boxes,
reflecting the discrepancy between the \textit{where} and the
\textit{what} concepts in exiting eye gaze
estimation algorithms. 

\begin{figure}
\begin{minipage}[c]{1.\linewidth}
    \centering
    \includegraphics[width=\linewidth]{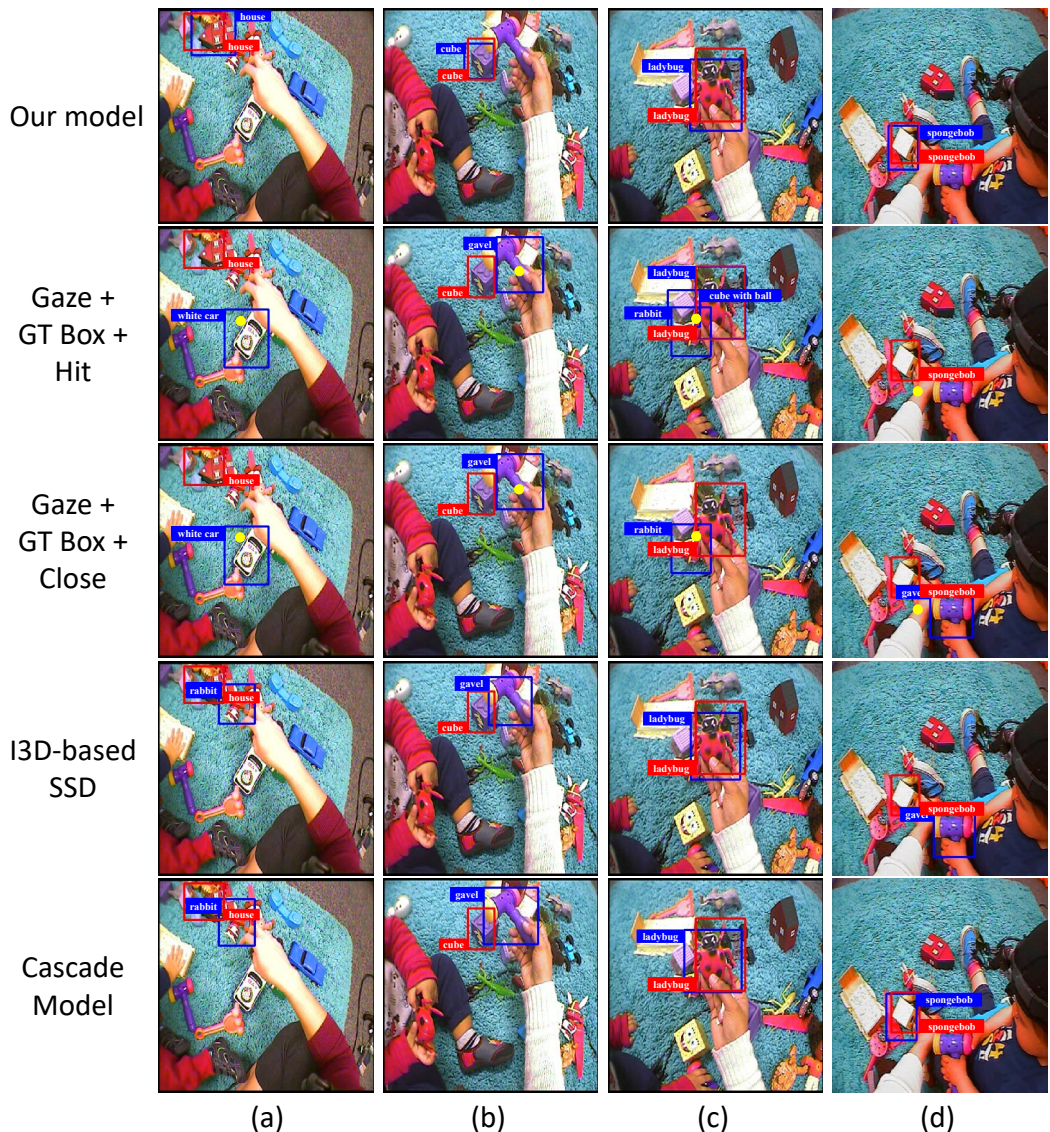}
    \caption{\emph{Sample results of our Mr.\ Net and baselines on ATT
        dataset.} Detections are in blue, ground truth  in red,
   and   the predicted gaze of gaze-based methods in yellow.}
    \label{fig:att}
\end{minipage} 
\vspace{-15pt}
\end{figure}

Sample results of our model compared with other baselines are shown in
Figure~\ref{fig:att}. Regular gaze prediction models fail in (c) \&
(d), supporting our hypothesis about the drawback of estimating where
and what independently --- the model is not robust to small errors in gaze estimation (recall the gaze-based baseline uses ground
truth bounding boxes so failures must be caused by gaze estimation).
In particular, the estimated gaze falls on 3 objects in (c), slightly
closer to the center of the rabbit; In (d), eye gaze does not fall on
any object. More unified models (I3D-based SSD, the
cascade model, and our model) thus achieve better performance. In (a)
\& (b), our model outperforms I3D-based SSD and Cascade. Because a
Self Validation Module is applied to inject consistency, our Mr. Net
performs better when many objects including the object of interest are
close to each other.

\begin{figure}
\begin{minipage}[t]{.32\linewidth}
    \centering
    \includegraphics[width=\linewidth]{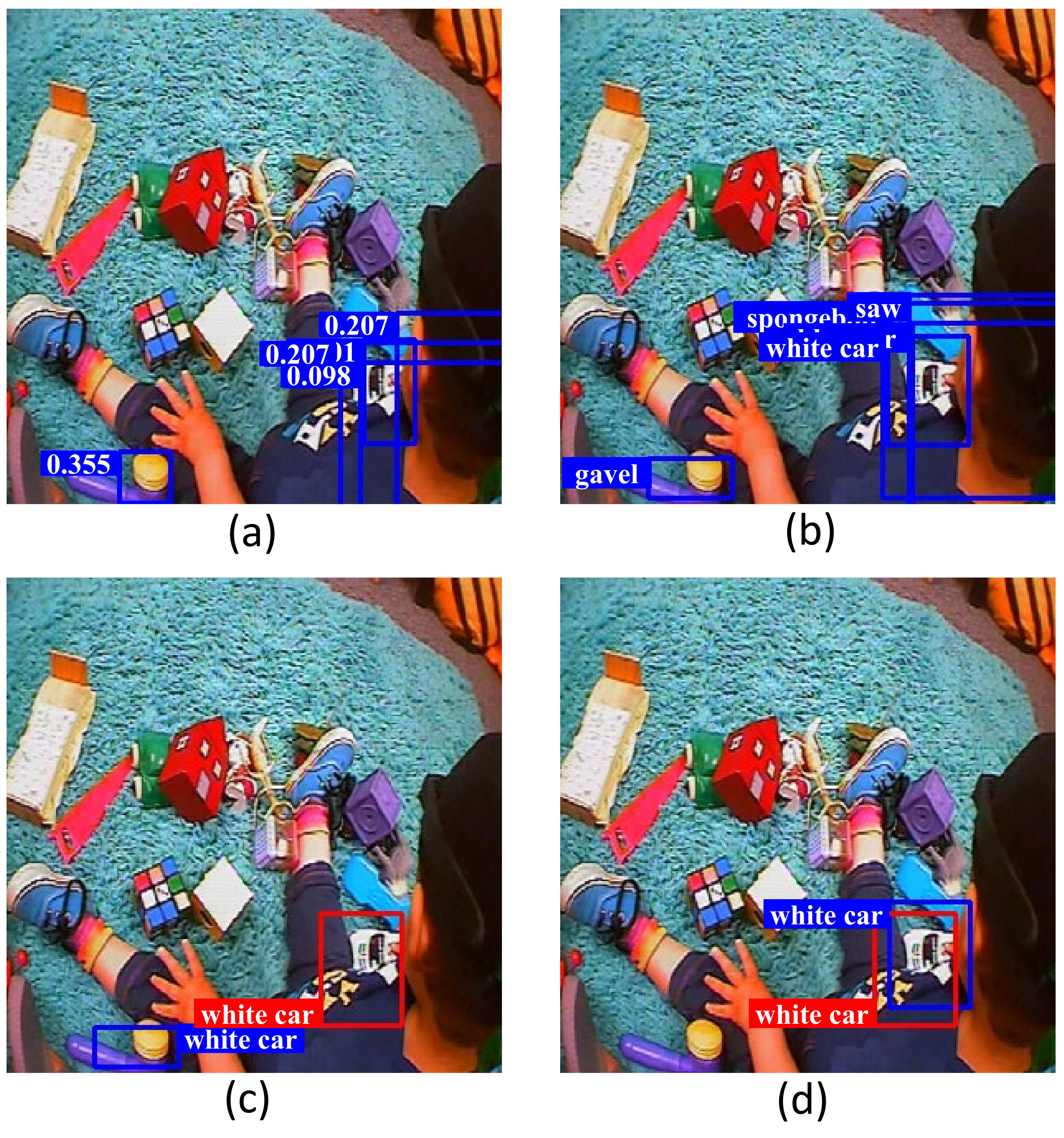}
    \caption{\textit{Illustration of how parts of our model work.}}
    \label{fig:parts}
\end{minipage}
\hfill
\begin{minipage}[t]{.32\linewidth}
    \centering
    \includegraphics[width=\linewidth]{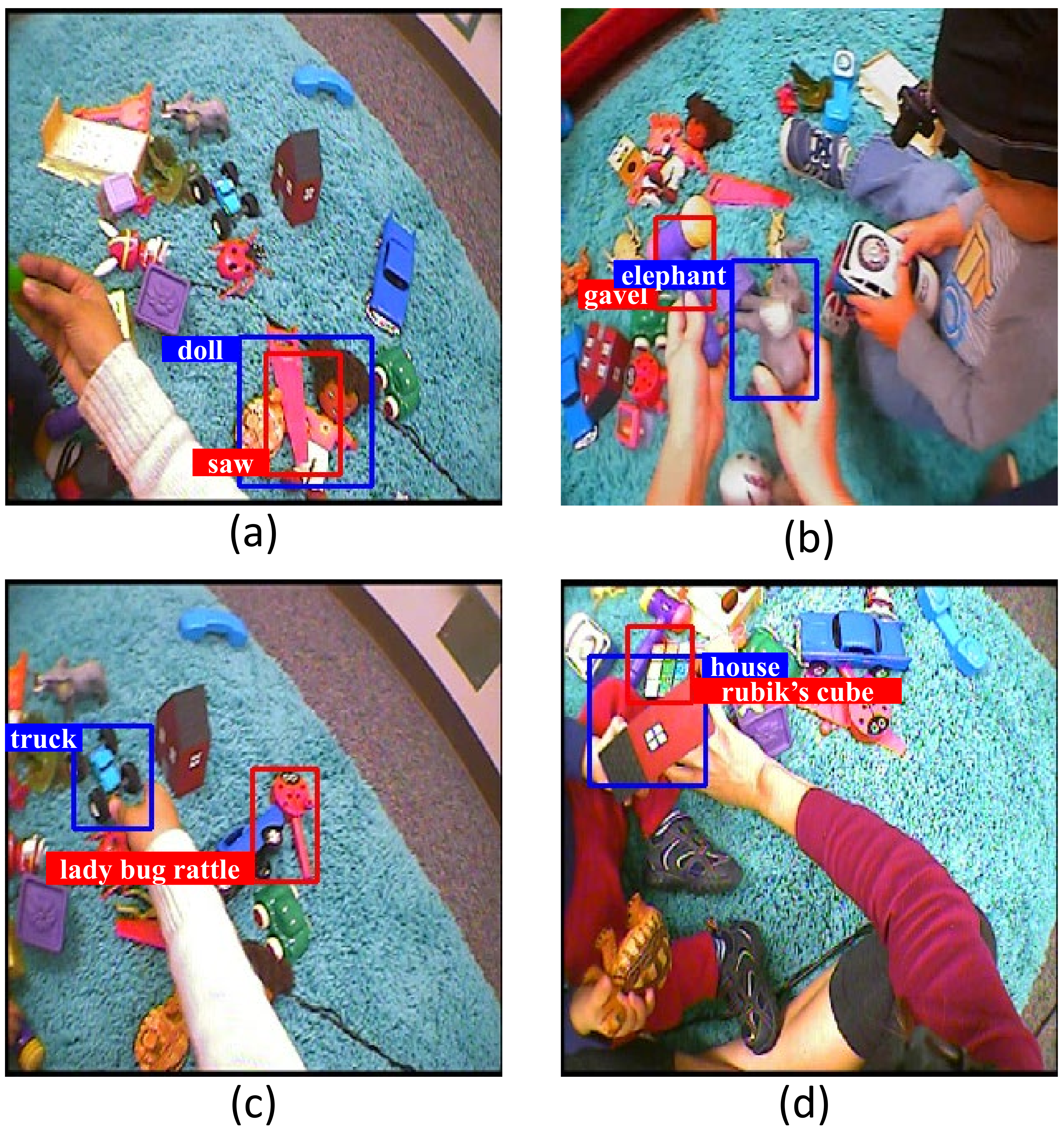}
    \caption{\emph{Some failure cases of our model,} with detections in blue and ground truth in red.}
    \label{fig:failure}
\end{minipage}
\hfill
\begin{minipage}[t]{.32\linewidth}
    \centering
   \includegraphics[width=\linewidth]{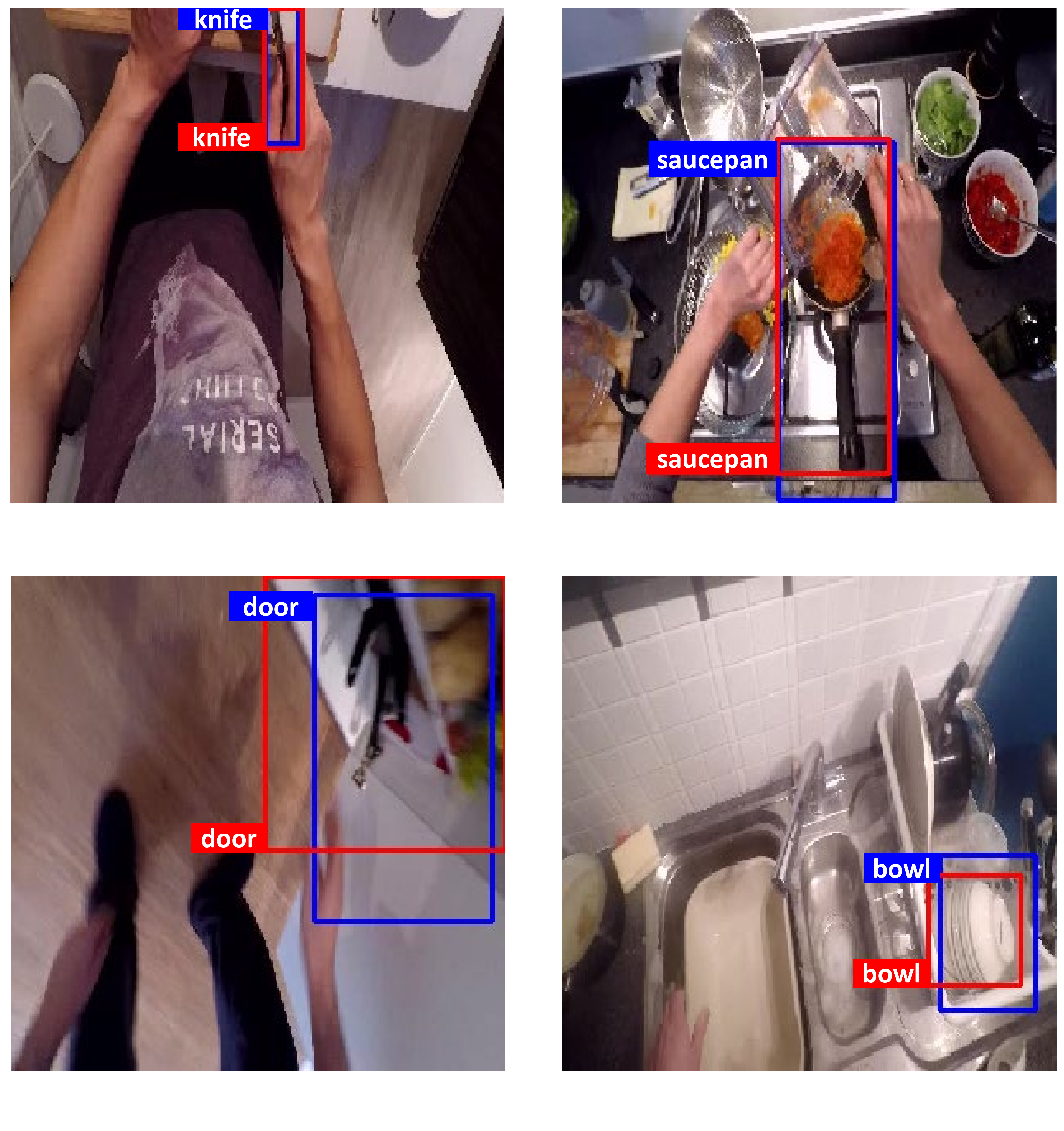}
   \caption{\textit{Sample results of Mr. Net on Epic-Kitchens.}}
   \label{fig:epic}
\end{minipage}
\vspace{-15pt}
\end{figure}

Figure~\ref{fig:parts} illustrate how various parts of our model work.
Image (a) shows the intermediate anchor attention score $A$$ \in$$
R^{a \times 1}$ from the temporal branch, visualized as the top $5$
attended anchors with attention scores. These are anchor-level
attention and no box offsets are predicted here. Image (b) shows
visualizations of the predicted anchor offsets $O$$ \in $$R^{a \times
  4}$ and box class score $C_{box} $$\in $$R^{a \times c}$ from the
spatial branch (only of the top 5 attended anchors). We do not have
negative samples or a  background class for training the
spatial branch and thus there are some false positives.  Image (c) combines
output from both branches; this is also the final prediction of the
model trained with the Self Validation Module but tested without it in
the ablation studies in Section~\ref{sec:svm}. The predicted class is
obtained from $C_{global}$ and we combine $A$ and $O$ to get the
location. Discrepancy happens in this example as the class prediction
is correct but not the location. Image (d) shows prediction of our full
model. By applying double self validation, the full model correctly
predicts location and class.

Some failure cases of our model are shown in
Figure~\ref{fig:failure}:
 (a) heavy occlusion, (b) ambiguity of which
held object is attended, (c) the model favors the object that is
reached for, and (d) an extremely difficult case where parent's reach
is occluded by an object held by the child.

\subsection{Ablation studies}

We conduct several ablation studies to evaluate the importance
of the parts of our model.

\textbf{Hard argmax vs. soft argmax during testing.}  The soft version
of what$\xrightarrow{}$where is necessary for gradient backpropagation
during training, but there is no such issue in testing. 
Our full
model achieves $mAcc=44.78\%$ when tested with hard argmax, versus
$mAcc=44.13\%$ when tested with soft argmax. When doing the same
experiments with other model settings, we observed similar results.

\textbf{Self Validation Module.}
\label{sec:svm}
To study the importance of the Self Validation Module, we conduct five
experiments: (1) Train and test the model without the Self
Validation Module; (2) Train the model without the Self Validation
Module but test with only the
what$\xrightarrow{}$where validation (the first step of Self Validation); 
(3) Train the model without Self Validation
but test with it; (4) Train the model with Self Validation
but test with only 
what$\xrightarrow{}$where validation; (5) Train the model with Self Validation
but test without it. As shown in Table~\ref{tab:ablation}, 
the Self Validation Module yields consistent performance
gain. If we train the model with Self Validation but remove
it during testing, the remaining model still outperforms other models
trained without the module.
This implies that embedding the Self Validation Module
during training helps learn a better model by bridging each component
and providing guidance of how components are related to each other. Even when  Self
Validation  is removed during testing, consistency is still
maintained between the temporal and the spatial branches. Also, recall
that when training the model with the Self Validation Module, the loss is
computed based on the final output, and thus when we test the full
model without Self Validation, the output is actually a
latent representation in our full model. This suggests that our Self
Validation Module encourages the model to learn a highly
semantically-meaningful latent representation. Furthermore, the
consistency injected by Self Validation helps prevent overfitting, 
while significant overfitting was observed without the Self Validation
Module during training.

\textbf{Validation method for what$\xrightarrow{}$where.}  We used
element-wise summation for what$\xrightarrow{}$where validation. Another
strategy is to treat $V_{attn}$ as an attention
vector in which rescaling is unnecessary,
\begin{equation}
\label{eq:softattnvali}
    A_i' = A_i \cdot \widetilde{V}^i_{attn}, 
    ~~~~~~~~~~~~~\mbox{with}~~~\widetilde{V}^i_{attn} = \frac{e^{V^i_{attn}}}{\sum_{j=1}^{a}e^{V^j_{attn}}}.
\end{equation}
We repeated experiments using this technique 
 and obtained $mAcc=43.30\%$, a slight drop 
that may be because the double softmax inside the Self Validation
Module increases optimization difficulty.

\textbf{Single stream versus two streams.}  We conducted experiments to study
the effect of each stream in our task. As
Table~\ref{tab:ablation} shows, a single optical flow stream performs much worse
than single RGB or two-stream, indicating that object appearance  is
very important for problems related to object detection. However,
it still acheived acceptable results since the network can refer to the
spatial branch for appearance information through the Self
Validation Module. To test this, we removed the Self
Validation Module from the single flow stream model during
training. When testing this model directly, we observed a very poor result
of $mAcc=18.4\%$; adding the Self Validation Module back during
testing yields a large gain to $mAcc=25.1\%$.

\textbf{Alternative matching strategy for box attention
  prediction.}\label{sec:onehot} For the anchor box attention
predictor, we perform experiments with different anchor matching
strategies. When multi anchor matching is used, we do hard negative
mining as suggested in~\cite{ssd} with the negative:positive ratio set
to 3. The model with the multi anchor
matching strategy achieves $mAcc=44.27\%$, versus $mAcc=44.78\%$ with one-best
anchor matching. We tried other different negative:positive ratios
(~\textit{e.g.} 5, 10, 20) and still found the one best anchor
matching strategy works better. This may be because we have an
acceptable number of anchor boxes; once we set more anchor boxes,
multi matching may work better.

\textbf{Object of interest class prediction.}  We explore where to
place the global object of interest class predictor. When we
connect it to the temporal branch after the fused block 5, we obtain $mAcc=44.78\%$;
when placed after the conv block 8 at the end of the temporal branch, we achieve
$mAcc=43.69\%$.
This implies that for
detecting the object of interest among others, a higher spatial
resolution of the feature map is helpful.

\begin{table}[t]
\begin{minipage}[t][][b]{.45\linewidth}
    \centering
    \medskip
\scalebox{0.7}{
\begin{tabular}{l c c c c } 
\toprule
         Model  & $Acc_{0.5}~\uparrow$ & $Acc_{0.75}~\uparrow$ & $mAcc~\uparrow$ \\ \midrule
        Mr. Net   & \textbf{71.34} & \textbf{38.26} & \textbf{39.04}\\ 
\midrule
        Gaze~\cite{t3f} + GT Boxes Hit & 26.46 & 26.46 & 26.46 \\
        Gaze~\cite{t3f} + GT Boxes Closest & 36.81 & 36.81 & 36.81 \\
        I3D~\cite{i3d}-bsaed SSD~\cite{ssd} & 67.43 & 37.90 & 37.22 \\
        Cascade Model & 65.96 & 38.01 & 37.93 \\
        \bottomrule
\end{tabular}}
\vspace{6pt}
      \caption{\textit{Results of online detection.}}
      \label{tab:online}
\end{minipage}\hfill
\begin{minipage}[t][][b]{.45\linewidth}
    \centering
    \medskip
\scalebox{0.7}{
\begin{tabular}{l c c c c c } 
\toprule
        Method  & $Acc_{0.5}~\uparrow$ & $Acc_{0.75}~\uparrow$ & $mAcc~\uparrow$ \\ \midrule
        Our Mr. Net   & \textbf{57.18} & \textbf{31.00} & \textbf{31.20}\\ 
\midrule
        I3D~\cite{i3d}-based SSD~\cite{ssd} & 47.58 & 24.38 & 25.42 \\
        Cascade Model & 51.20 & 28.18 & 28.36 \\
        \bottomrule
\end{tabular}}
\vspace{6pt}
      \caption{\textit{Accuracies on the   Epic-Kitchen dataset.}}
      \label{tab:epic}
\end{minipage}
\vspace{-20pt}
\end{table}

\subsection{Online Detection}
Our model can be easily modified to do online detection, in which only
previous frames are available. We modified
the model to detect the object of interest in the
last frame of a given sequence. As shown in
Table~\ref{tab:online}, except for the Gaze + GT boxes model, all other
models suffer from dropping $Acc$ scores, indicating that online
detection is more difficult. However, since the gaze prediction model
that we use~\cite{t3f} is trained to predict eye gaze in each frame of the
video sequence and thus works for both online and offline tasks, its
performance remains stable.

\subsection{Results on Epic-Kitchen Dataset}
We show the generalizability of our model by performing experiments on 
Epic-Kitchens~\cite{epickit}. Results by applying our model as
well as the I3D-based SSD model and the cascade model on this dataset
are shown in Table~\ref{tab:epic}. On this dataset,
the $Acc_{0.5}$ of the Cascade model is higher than that of the I3D +
SSD model. The reason may be that objects are sparser in this dataset
and thus poorly-predicted boxes will be less likely to lead to wrong
classification. Sample results are shown in Figure~\ref{fig:epic}.

\section{Conclusion}

We considered the problem of detecting attended object in cluttered
first-person views.
We proposed a novel unified model with a Self Validation Module to
leverage the visual consistency of human vision system.
The module jointly optimizes the class  and the
attention estimates as self
validation. Experiments on two public datasets show our model
outperforms other state-of-the-art methods by a large margin.

\section{Acknowledgements}

This work was supported in part by the
National Science Foundation (CAREER IIS-1253549), the National
Institutes of Health (R01 HD074601, R01 HD093792), NVidia, Google, and the IU
Office of the Vice Provost for Research, the College of Arts and
Sciences, and the School of Informatics, Computing, and Engineering
through the Emerging Areas of Research Project ``Learning: Brains,
Machines, and Children.''

{\small
\bibliographystyle{ieee}
\bibliography{mybib}
}

\pagebreak
\section{Supplementary Material}

\subsection{The architecture of the Cogged Spatial-Temporal Module}

\begin{figure*}[h]
    \begin{center}
       \includegraphics[width=1.0\textwidth]{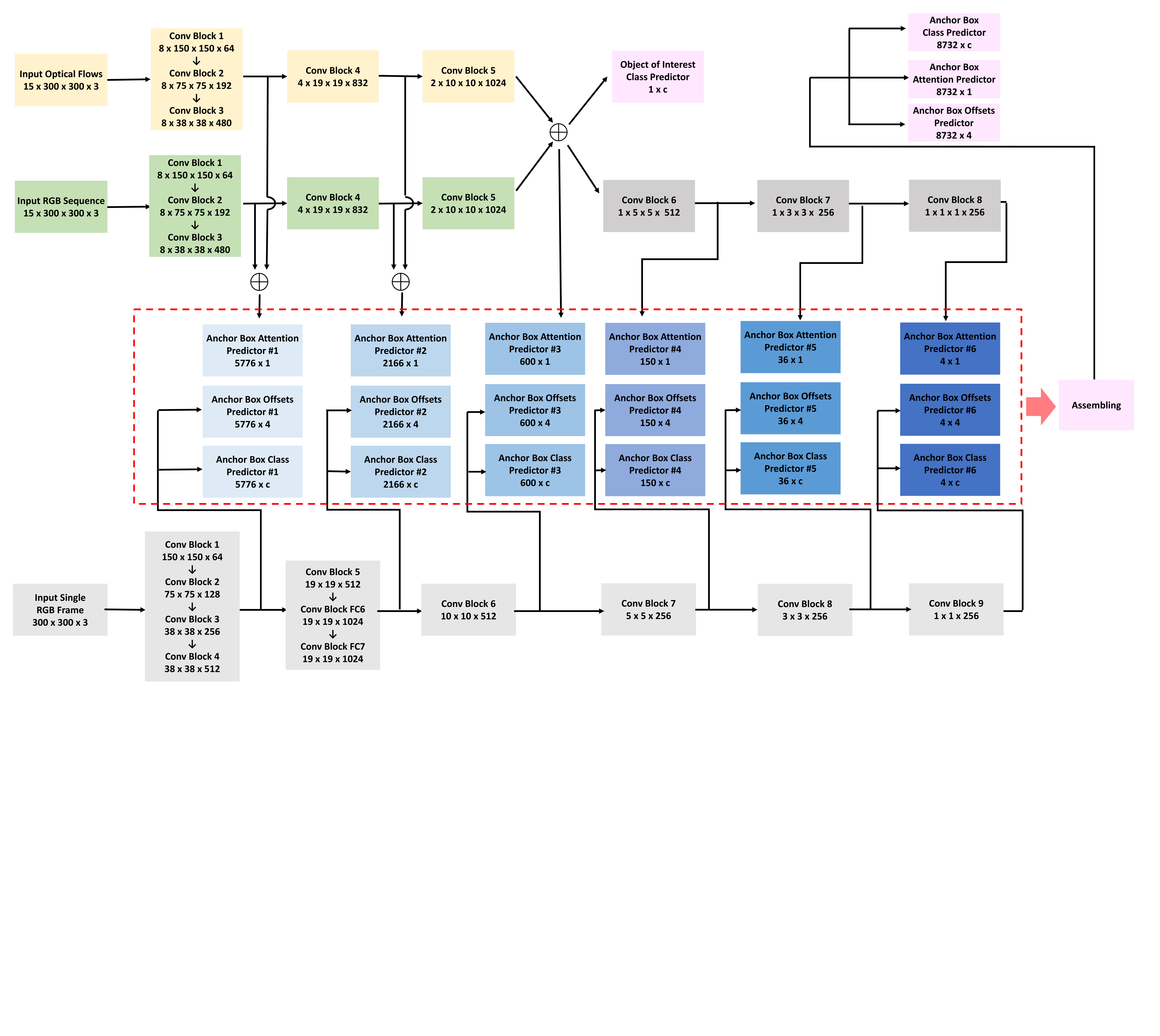}
    \end{center}
    \caption{\emph{The architecture of the Cogged Spatial-Temporal Module.} The number below each component indicates its output dimension. $c$ is the number of class. All fusion is performed by element-wise sum. When trained without being followed by the Self Validation Module, before computing $L_{globalclass}$, $L_{boxclass}$ and $L_{attn}$, Softmax is applied (the attention prediction is first flattened to be a 8732-d vector)}
    \label{fig:cogs}
\end{figure*}

\begin{figure*}[h]
    \begin{center}
       \includegraphics[width=1.0\textwidth]{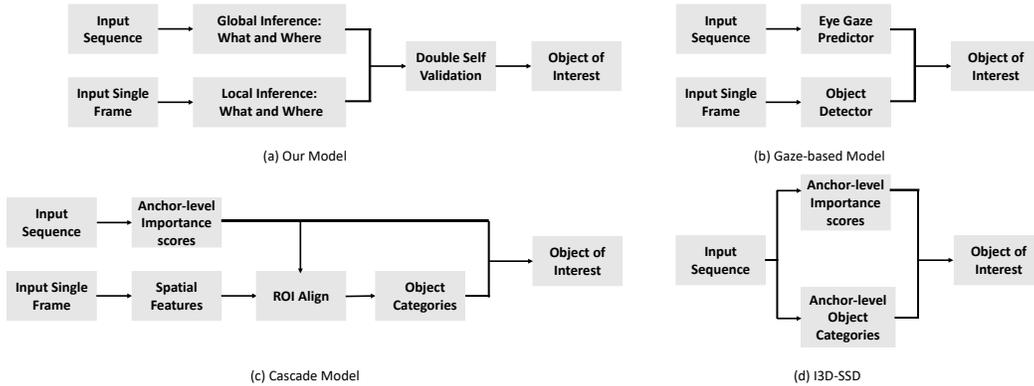}
    \end{center}
    \caption{\textit{Visualizations of (a) Our model, (b) Gaze-based model, (c) Cascade Model, and (d) I3D-backboned SSD.} Note that in our experiments of Gaze + Box model, we directly use ground truth bounding boxes for each object instead of results from an object detector. The box regression head is omitted for simplicity.}
    \label{fig:baselines}
\end{figure*}

\subsection{Hand based model settings}

We train two object-in-hand detectors (for the left hand and the right hand respectively), using the ResNet-50 backbone, and one which-hand classifier with the I3D backbone to classify which hand holds the object of interest when the left hand and the right hand hold different objects. During testing, if only one object-in-hand detector predicts object in hand or both hands hold the same object, we accept the prediction as the object of interest and it is combined with the ground truth bounding box as the final output. Otherwise we apply the which-hand classifier to decide which object to take. We obtain testing accuracy of $86.28\%$, $88.61\%$ and $90.80\%$ for the object-in-left-hand detector, the object-in-right-hand-detector and the object-in-which-hand classifier respectively.

To further strengthen the baseline, we directly use the ground truth of objects in hands and have 4 more settings: (1) Right handed model, which uses the ground truth object in hands labels, and when two hands hold different objects, it always favours the right one; (2) Left handed model, which is the same as (1) but always favours the left hand; (3) Model with object-in-hand ground truth and which-hand classifier, which will apply the which-hand classifier to decide which object to take when two hands hold different objects; (4) Either handed model, which uses the ground truth object-in-hand labels, and when two hands hold different objects, the model always take the one resulting in higher $mAcc$ as the prediction. Note that (4) depicts the best performance which hand-based methods can possibly achieve in theory as it uses all of the ground truth.

\begin{figure*}[h]
    \begin{center}
       \includegraphics[width=1.0\textwidth]{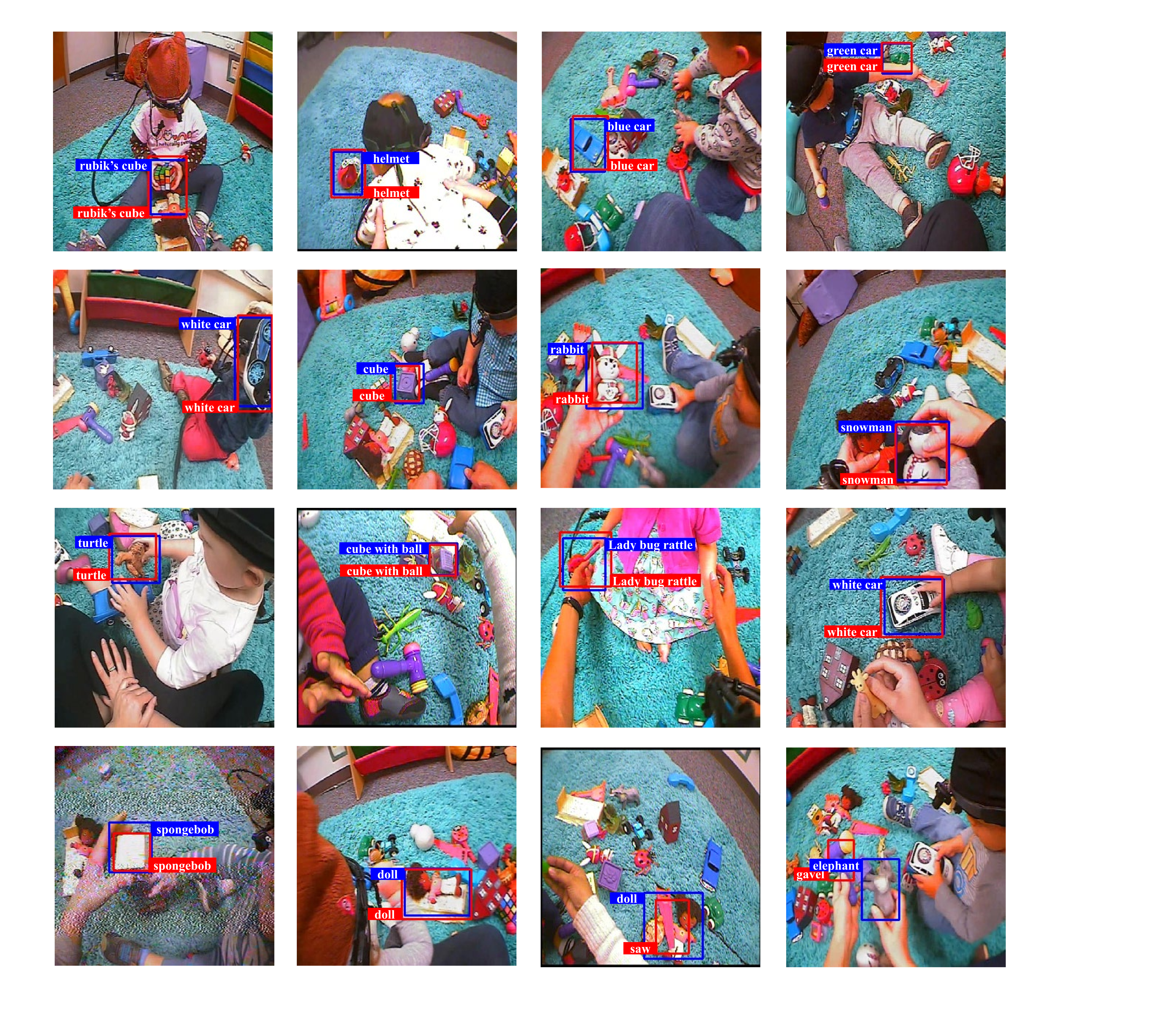}
    \end{center}
    \caption{\emph{More qualitative results of our model on the ATT dataset.}}
    \label{fig:moreatt}
\end{figure*}

\end{document}